\documentclass[letterpaper, 10 pt, conference]{ieeeconf}  % Comment this line out if you need a4paper

\IEEEoverridecommandlockouts                              % This command is only needed if 
\usepackage{epsfig}
\usepackage{graphicx}
\usepackage{amsmath}
\usepackage{amssymb}

\usepackage{dsfont}
\usepackage{siunitx}
\usepackage{subfigure}
\usepackage{multirow}
\usepackage{booktabs} % professional-quality tables
\usepackage{standalone}
\usepackage{xspace}
\usepackage{comment}
\usepackage{multirow}
\usepackage{color}
\usepackage{pifont}
\usepackage{xfrac}

\usepackage[pagebackref=true,breaklinks=true,letterpaper=true,colorlinks,bookmarks=false]{hyperref}

\newcommand{\myparagraph}[1]{\vspace{3pt}\noindent{\bf #1}}

\def\ours{StopNet\xspace}

\newcommand{\ie}[1][ ]{{\em i.\thinspace{}e\@.{}},#1}
\newcommand{\eg}[1][ ]{{\em e.\thinspace{}g\@.{}},#1}

\definecolor{col1}{RGB}{213, 229, 255}
\definecolor{col2}{RGB}{128, 179, 255}
\definecolor{col3}{RGB}{181, 232, 151}
\definecolor{col4}{RGB}{235, 155, 221}

\title{\LARGE \bf
\ours: Scalable Trajectory and Occupancy Prediction \\for Urban Autonomous Driving
}

\author{Jinkyu Kim$^{*1}$, Reza Mahjourian$^{*2}$, Scott Ettinger$^{2}$, Mayank Bansal$^{2}$,\\
Brandyn White$^{2}$, Ben Sapp$^{2}$, and Dragomir Anguelov$^{2}$% <-this % 
\thanks{$^{*}$Equal contribution. Correspondence: {\tt\small rezama@waymo.com}}%
\thanks{$^{1}$Department of Computer Science and Engineering, Korea University, Seoul 02841, South Korea. Work done while at Waymo.}
\thanks{$^{2}$Waymo, Mountain View, CA 94043, USA.}
}

\begin{document}

\maketitle
\thispagestyle{empty}
\pagestyle{empty}

\begin{abstract}
We introduce a motion forecasting (behavior prediction) method that meets the latency requirements for autonomous driving in dense urban environments without sacrificing accuracy. A whole-scene sparse input representation allows \ours to scale to predicting trajectories for hundreds of road agents with reliable latency. In addition to predicting trajectories, our scene encoder lends itself to predicting whole-scene probabilistic occupancy grids, a complementary output representation suitable for busy urban environments.  Occupancy grids allow the AV to reason collectively about the behavior of groups of agents without processing their individual trajectories. We demonstrate the effectiveness of our sparse input representation and our model in terms of computation and accuracy over three datasets. We further show that co-training consistent trajectory and occupancy predictions improves upon state-of-the-art performance under standard metrics.
\end{abstract}

\section{Introduction}
\noindent An Autonomous Vehicles (AV) needs to continuously evaluate the space of all possible future motions from other road agents so that it can maintain a safe and effective motion plan for itself.  This motion forecasting and re-planning task is one of the many processes that are continuously executed by the AV, so it is critical that it completes under expected latency requirements.  On the other hand, operating in dense urban environments, the AV may encounter scenes with hundreds of dynamic agents within its field of view---consider driving next to a sports or music venue with lots of pedestrians.  Autonomous driving in such environments requires a motion forecasting and planning system that is \ding{192} fast, \ding{193} scales well with the number of agents.

The existing motion forecasting methods do not meet the requirements discussed above.  Models typically take upwards of 40-50ms for inference.  This scalability issue is not addressed in public benchmarks~\cite{chang2019argoverse, interactiondataset, caesar2020nuscenes, lyft2020} and is often ignored in publications.  Proposed methods often use \emph{raster} (render-based) input representations~\cite{casas2018intentnet,hong2019rules, chai2019multipath,bansal2018chauffeurnet}, which require costly CNNs for processing.  Recently, methods have been proposed that use \emph{sparse} point-based input representations~\cite{gao2020vectornet, zhao2020tnt, khandelwal2020if, tolstaya2021identifying}.  These methods offer improvements in accuracy and a reduction in the number of model parameters.  However, with a focus on accuracy, these methods use \emph{agent-centric} scene representations, which require re-encoding road points and agent points from the view point of each individual agent.  The latency of these methods grows linearly with the number of inference agents, so they are not suitable for busy urban environments.

\begin{figure}[!t]
    \centering
        \includegraphics[width=\linewidth]{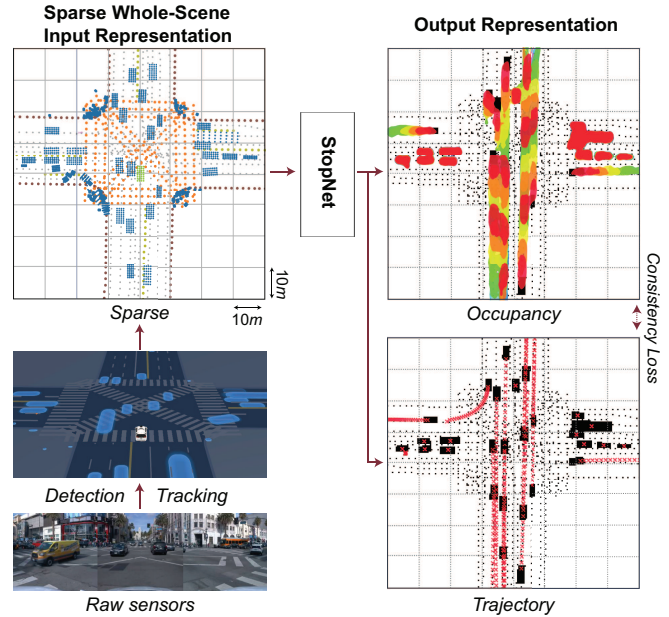}
    \caption{\ours uses a whole-scene sparse input representation, supporting a scalable motion forecasting model that unifies occupancy grids and trajectories.}
    \label{fig:teaser} \vspace{-0.5em}
\vspace{-10pt}
\end{figure}

This work introduces \ours, a motion forecasting method focused on latency and scalability.  We develop a novel \emph{whole-scene} sparse input representation which can encode scene inputs pertaining to all agents at once.  Drawing from the 3D object detection literature, we develop a PointPillars-inspired~\cite{lang2019pointpillars} scene encoder to concurrently process sparse points sampled from all agents, leading to a very fast trajectory prediction model whose latency is mostly invariant to the number of agents.

The predicted trajectories and uncertainties are often consumed as planning constraints by the AV, therefore the latency of the planning algorithm also increases in busy scenes.  \ours's whole-scene encoder also supports predicting probabilistic occupancy grids~\cite{thrun1996integrating}---a dense output format capturing the probability that any given grid cell in the map is occupied by some agent part.  This output representation allows the AV planner to reason about the occupancy of entire regions in busy scenes without a need for processing individual trajectories---thereby requiring almost constant computation.  Another attractive property of occupancy grids is that they are robust to detection and tracking noise and flicker, since they allow the model to infer occupancy independently of agent identity over time.

Via a co-training setup, \ours is also the first method to unify trajectory sets and occupancy grids as the two archetypes of motion forecasting.  We tie together these output representations with an intuitive consistency loss: the per-agent trajectory output distribution, when converted to an occupancy probability distribution, should agree with the overall occupancy distribution. Our experiments show that co-training in this manner leads to state-of-the-art trajectory prediction performance.

\section{Related Work}
\myparagraph{Agent-Centric vs. Whole-Scene Modeling.}
While there are other alternatives, most prediction methods rely on a sequence of agent state observations often provided by a detection/tracking system~\cite{rudenko2020bp_survey}.  Agent-centric models re-encode the world from the view point of every agent in the scene~\cite{gao2020vectornet, zhao2020tnt, tang2019multiple, rhinehart2018r2p2, hong2019rules, khandelwal2020if, mercat2020multi, tolstaya2021identifying, zhao2019multi, sriram2020smart}.  This process requires transforming road state and the state of all other agents into an agent-centric frame. Therefore, these methods scale linearly with the number of agents, which poses a scalability issue in dense urban scenes with hundreds of pedestrians and vehicles. A popular alternative is whole-scene modeling~\cite{casas2018intentnet, bansal2018chauffeurnet, chai2019multipath, phan2020covernet, biktairov2020prank, buhet2020plop}, where the bulk of the scene encoding is done in a shared coordinate system for all agents.  Whole-scene modeling has the very attractive advantage that the processing time is invariant to the number of agents.

\myparagraph{Dense vs. Sparse Input Representation.}
To our knowledge, whole-scene models have always used a bird's-eye view (BEV) raster input representation to encode road elements, agent state, and agent interactions. This approach allows including a variety of heterogeneous inputs into a common raster format, and enables the use of well-established powerful CNN models.  However, there are several disadvantages.  The model's field of view (FOV) and resolution are constrained by the computational budget, and the ability to model spatially-distant interactions is dependent on the receptive field of the network.  Finally, while it is possible to render some state attributes, \eg vehicle extent, it is unclear how to rasterize some attributes, like uncertainty over agent orientation.  On the other hand, with sparse inputs representations~\cite{khandelwal2020if, zhao2020tnt, gao2020vectornet, tolstaya2021identifying} the model inputs consist of vectors of continuous state attributes encoding the agent motion history, relation to road elements, and relation to neighboring agents.  This allows for arbitrary long-range interactions, and infinite resolution in continuous state attributes.  However, sparse inputs have always been combined with agent-centric models, posing scalability issues.  \ours is the first method to address scalability by introducing a whole-scene sparse input representation and model.

\myparagraph{Trajectory vs. Occupancy Output Representation.} Representing future motion is traditionally done in two ways. The popular approach is a parametric distribution over a set of trajectories per agent~\cite{casas2018intentnet, gao2020vectornet, tang2019multiple, rhinehart2018r2p2, khandelwal2020if, mercat2020multi, chai2019multipath, phan2020covernet, biktairov2020prank, buhet2020plop}. A common approach to capturing trajectory uncertainty is to predict multiple trajectories per agent as well as Gaussian position uncertainty for each trajectory waypoint, which in busy scenes, amounts to a large set of constraints to process in the planning algorithm.   Moreover, the per-agent trajectories may be overlapping in space, and sampling from them independently may produce samples which violate physical occupancy constraints by placing agents on top of each other.
An alternative output representation is to predict the collective occupancy likelihood as discretized space-time cells in a grid view of the world~\cite{hong2019rules, jain2020discrete, bansal2018chauffeurnet, djuric2020multinet, strohbeckmultiple, casas2021mp3}.  While occupancy grid models have been mentioned in passing~\cite{bansal2018chauffeurnet} and embedded in other tasks~\cite{casas2021mp3}, in this work we study them in detail and develop metrics to evaluate them.

\section{Method}\label{sec:model}

\begin{figure*}[!t]
    \vspace{0.5em}
    \centering
        \includegraphics[width=\linewidth]{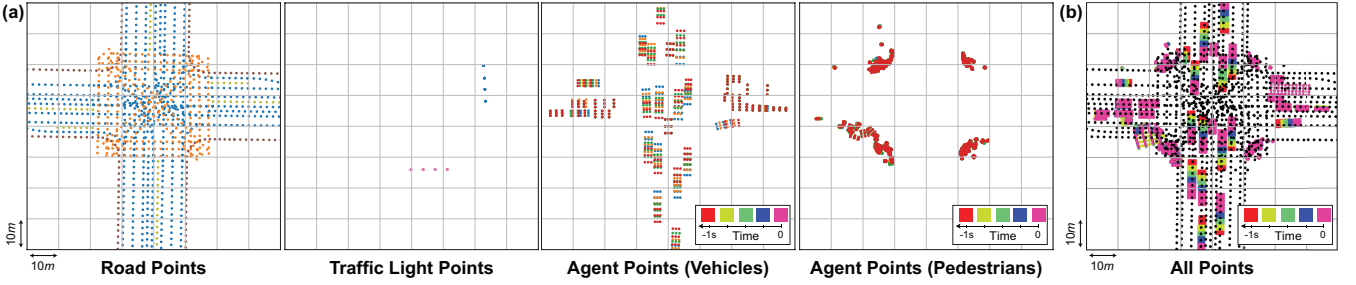}
    \caption{Sparse Whole-Scene Input Representation. (a) Input point sets $\mathcal{P}^{r}$, $\mathcal{P}^l$ and $\mathcal{P}^{a}$ (vehicles and pedestrians) for an example scene. (b) All points.}
    \label{fig:repr}\vspace{-.5em}
\end{figure*}

\subsection{Problem Definition}\label{sec:problem}
We assume that each agent at any time $t$ can be represented by an oriented box as a tuple ($\mathbf{s}_t$, $\theta_t$, $w_t$, $l_t$, $v_t$, $a_t$), where $\mathbf{s}_t = (x_t, y_t)$ denotes the agent's 2D center position, $\theta_t$ denotes the orientation, ($w_t$, $l_t$) denote box extents, and $v_t$, $a_t$ denote 2D velocity and acceleration vectors.  Given a sequence of state observations over a fixed number of input timesteps for all agents in the scene, the \textbf{Trajectory Prediction} task is defined as predicting the future positions $\hat{\mathbf{s}}_t$, $t \in \{1,\dots,T\}$ for all agents in the scene over a fixed time horizon $T$.  Following previous methods~\cite{chang2019argoverse, chai2019multipath}, we predict a set of $K$ trajectories $\hat{\mathbf{s}}_t^k$, $k \in \{1,\dots,K\}$ with associated probabilities for each agent.  We also predict 2D Gaussian uncertainties for each trajectory waypoint $\hat{s}_t^k$.

The \textbf{Occupancy Prediction} task is defined as predicting occupancy grids $\hat{O}_t$, $t \in \{1,\dots,T\}$ with spatial dimensions $W \times H$. Each cell $\hat{O}_t(x, y)$ in the occupancy grid $\hat{O}_t$ contains a value in the range $[0, 1]$ representing the probability that any part of any agent box overlaps with that grid cell at time $t$.  The ground-truth occupancy grids are constructed by rendering future agent boxes in BEV as binary maps.  Since the planner reacts to different agent classes differently, we predict separate occupancy grids for each agent class.

\subsection{Sparse Whole-Scene Input Representation}~\label{ss:sparse-input}
We use a whole-scene coordinate system centered on the AV's position at $t = 0$ (see Fig.~\ref{fig:repr}).  All the current and past agent states (including the AV's) are transformed to this fixed coordinate system.  The model inputs consist of three sets of points $\mathcal{P} = \mathcal{P}^r \cup \mathcal{P}^l \cup \mathcal{P}^a$, each with associated feature vectors.  Agent points $\mathcal{P}^a$ are constructed by uniformly sampling a fixed number of points from the interior of each agent box.  The agent points from all input timesteps co-exist.  Each agent point carries the state attributes mentioned in Sec.~\ref{sec:problem}, plus a one-hot encoding of time.  The road element points $\mathcal{P}^r$ are sampled uniformly from the lines and curves of the road structure.  Each road point encodes position and element type.  Traffic light points $\mathcal{P}^l$ are placed at the end of the traffic lanes that they control.  Their attributes include position, time, and traffic light state (color).

\subsection{Whole-Scene Encoder}~\label{ss:encoder}
\begin{figure*}[!t]
    \centering
        \includegraphics[width=\linewidth]{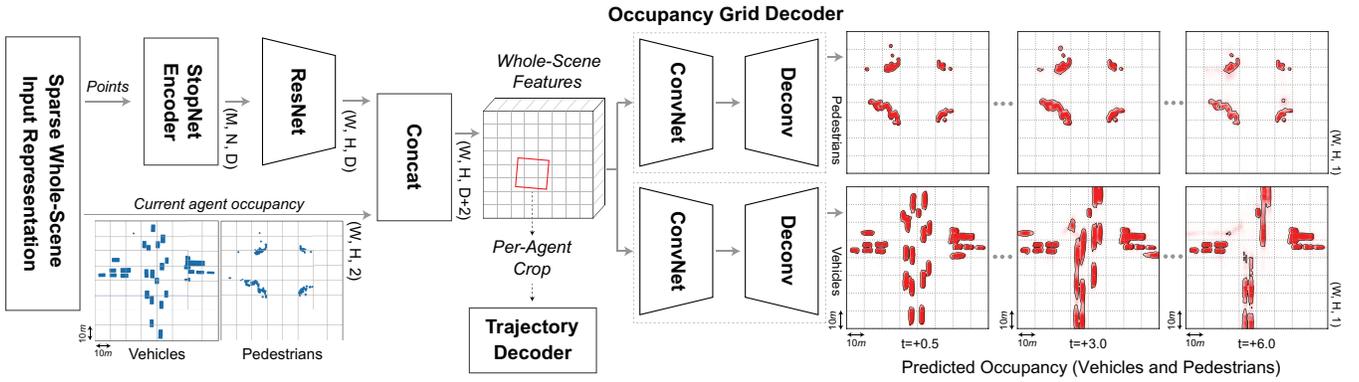}
    \caption{An overview of the \ours architecture. The encoder processes the input point set $\mathcal{P}$ and produces a feature map, which is used to predict both per-agent trajectories and whole-scene occupancy grids for each agent type. Input agent boxes at $t=0$ are also rendered in BEV as binary features and fed to the trajectory and occupancy grid decoders.}
    \label{fig:arch}
    %\vspace{-1em}
\end{figure*}

Fig.~\ref{fig:arch} shows an overview of the \ours architecture. It consists of an encoder, a ResNet backbone, and two heads for decoding trajectory and occupancy predictions from the shared scene features.

Inspired by PointPillars~\cite{lang2019pointpillars}, the \ours encoder discretizes the point set $\mathcal{P}$ into an evenly-spaced grid of  $M$$\times$$N$ pillars in the x-y plane, $\{\pi_1,\pi_2,\ldots,\pi_{MN}\}$. The points in each pillar are then augmented with a tuple ($x_c$, $y_c$, $x_\textnormal{offset}$, $y_\textnormal{offset}$) where the $c$ subscript denotes distance to the arithmetic mean of all points in the pillar and the \emph{offset} subscript denotes the offset from the pillar center. We then apply a simplified version of PointNet~\cite{qi2017pointnet} to encode and aggregate the features from all points in each pillar $\pi_j$.  In particular, we apply a linear fully-connected (FC) layer followed by BatchNorm and a ReLU to encode each point. A max operation is then applied across all the points within each pillar to compute a single feature vector per pillar as
\begin{equation}
    f_{\pi_j} = \textnormal{MaxPool}\Bigl(\{\textnormal{ReLU}(\textnormal{BN}(\textnormal{FC}(p_i)))\}_{p_i\in\pi_j}\Bigr).
\end{equation}

The $M \times N$ feature map produced by the encoder is then processed through a ResNet backbone, reshaped to $W \times H$, and concatenated with binary occupancy grids rendered from the current positions of scene agents.  The resulting feature map is then shared by a trajectory decoder and an occupancy grid decoder to produce the final predictions of the model.

\subsection{Per-Agent Trajectory Decoder}\label{ss:co-train}
To predict trajectories, we use the trajectory decoder architecture and losses from MultiPath~\cite{chai2019multipath}.  The trajectory decoder extracts patches of size $11\times 11$ centered on each agent location from the whole-scene features, thus operating on a per-agent basis. Note that while trajectory prediction head is agent-centric, the bulk of the model computation is whole-scene, and this dominates the overall processing time.

The trajectory decoder uses a fixed set of pre-clustered potential trajectories as an \emph{anchor-set}, and ground-truth trajectories are assigned an anchor via closest Euclidean distance.  For each anchor, the decoder regresses per-waypoint deltas from the anchor trajectory, yielding a Gaussian mixture at each timestep.  The losses consist of a softmax cross-entropy classification loss over anchors $\mathcal{L}_{s}$, and within-anchor squared $L_{2}$-norm regression loss $\mathcal{L}_{r}$.

\subsection{Occupancy Grid Decoder}\label{ss:heatmap-decoder}
The occupancy grid decoder processes the whole-scene feature map at once through a very lightweight CNN, which is repeated for each timestep $t$ and produces occupancy logits for each class $a$ as separate channels.  The per-cell occupancy probabilities are obtained by applying a sigmoid function to the logits.  The occupancy loss is defined as
\begin{equation}
     \mathcal{L}_{o}(\mathcal{\hat{O}}, \mathcal{O}) = \frac{1}{WH}\sum_a\sum_{t}\sum_{x}\sum_{y}\mathcal{H}(\mathcal{\hat{O}}^{a}_{t}, \mathcal{O}^{a}_{t}),
\end{equation}
where $\mathcal{H}$ denotes the cross-entropy function and $\mathcal{O}^{a}_{t}$ denotes the ground-truth occupancy for agent class $a$ at time $t$.

\subsection{Co-Training and Consistency Loss}
In addition to co-training the trajectory and occupancy decoders, we find it useful to employ a consistency loss to encourage agreement between the per-agent trajectory predictions and whole-scene occupancy grids.  The trajectory predictions with the highest predicted likelihood are rendered as oriented bounding boxes and aggregated by agent class as $\mathcal{\tilde{O}}_{t}^{a}$.  Consistency with predicted occupancy outputs $\mathcal{\hat{O}}^{a}_{t}$ is then computed similarly to computing cross-entropy with the ground-truth as $\mathcal{L}_{c}(\mathcal{\hat{O}}, \mathcal{\tilde{O}}) = \mathcal{L}_{o}(\mathcal{\hat{O}}, \mathcal{\tilde{O}})$.

The loss function for the most general variant of our model is then summarized as
\begin{equation}
    \mathcal{L} = \underbrace{\lambda_{o}\mathcal{L}_{o}}_{\parbox{4em}{\scriptsize\text{Occupancy Loss}}} + \overbrace{\lambda_{s}\mathcal{L}_{s} + \lambda_{r}\mathcal{L}_{r}}^{\text{Trajectory Loss}} + \underbrace{\lambda_{c}\mathcal{L}_{c}}_{\parbox{2em}{\scriptsize\text{Consistency Loss}}}
\end{equation}
where $\lambda_{o}$, $\lambda_{s}$, $\lambda_{r}$, and $\lambda_{c}$ are the respective loss weights.

\begin{figure*}[!t]
    \centering
        \includegraphics[width=.95\linewidth]{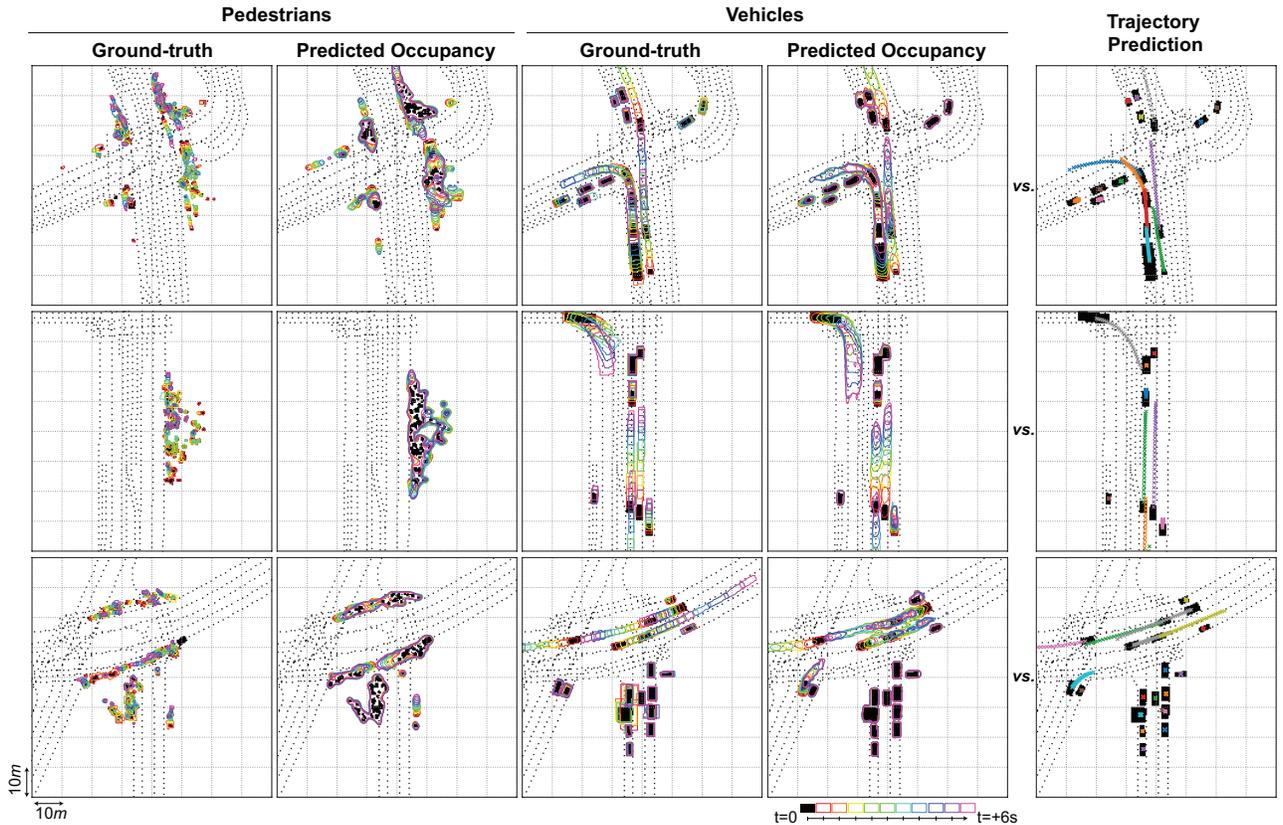}
    \caption{Example occupancy and trajectory predictions from \ours.  \textbf{Left four columns:} Ground-truth and predicted occupancy grids are visualized through time as color-coded contour lines (from red for near future to purple for far future), where each contour contains values with probability $>$ 0.5.  \textbf{Right column:} For trajectories, different colors map to different agents.  The dotted lines represent road points and the black boxes represent the current location of agents at $t=0s$.}
    \label{fig:viz-examples} \vspace{-1em}
\end{figure*}

\section{Experiments}
\subsection{Datasets}
\myparagraph{Crowds Dataset.}
This dataset is a revision of the Waymo Open Motion Dataset~\cite{ettinger2021large} focused on crowded scenes.  It contains over 13 million scenarios spanning over 500 hours of real-world driving in several urban areas across the US.  The scenarios contain dynamic agents, traffic lights and road network information.  All scenarios contain at least 20 dynamic agents.

\myparagraph{Interaction \& Argoverse Datasets.}
We also evaluate our proposed method on the Interaction~\cite{interactiondataset} and  Argoverse~\cite{chang2019argoverse} datasets. %These datasets include contextual information such as trajectory histories, context agents, and lane centerlines. The Interaction dataset provides different types of interactive driving scenarios, making this dataset valuable for the prediction problem due to its emphasis on vehicle-to-vehicle interactions. The Argoverse~\cite{chang2019argoverse} dataset poses the forecasting task for only a single agent, while all other dynamic objects are considered as context agents. This makes this dataset less interesting for our approach as our focus is on multi-agent prediction in dense urban areas. We, however, report scores on this dataset.
The Interaction dataset contains interactive driving scenarios involving multiple agents.  In the Argoverse dataset, only one agent has future ground-truth, making it less interesting for our multi-agent whole-scene method.  We, however, report scores on this dataset as well.

\subsection{Training Setup}
% \myparagraph{Model Details.}
We train three variants of our model: $M_T$ is trained only with a trajectory loss, $M_O$ is trained only with an occupancy loss, and $M_{TO}$, which uses co-training and a consistency loss.  All models are trained from scratch using an Adam optimizer~\cite{kingma2014adam}, with a learning rate of $0.0004$ and batch size of $8$.  We clip the gradient norms~\cite{pascanu2013difficulty} above $0.1$.  The loss weights are $\lambda_o=100.0$, $\lambda_s=1.0$, $\lambda_r=0.16$, and $\lambda_c=10.0$, determined using light grid search.  The input field of view is $160m$$\times$$160m$, corresponding to an effective sensing range of $80m$ for the AV.
Our encoder uses $M$$\times$$N$ = $80$$\times$$80$ pillars.  We sample $8 \times 8$ input points uniformly from the interior of all agent boxes.  Our occupancy decoder has a resolution of $W$$\times$$H$ = $400$$\times$$400$, predicting occupancy over $T = 10$ linearly-spaced timesteps up to 6 seconds in the future, \ie $t\in\{0.6, 1.2, \dots, 6.0\}$. All figures show an $80m\times80m$ center crop of the output to show more details.

\begin{figure}[!t]
    \centering
        \includegraphics[width=\linewidth]{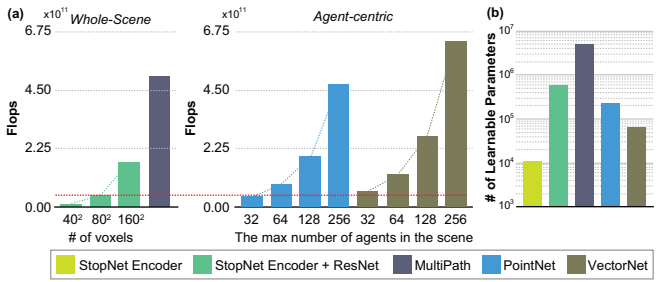}
    \caption{Comparison of \textbf{(a)} number of flops, and \textbf{(b)} number of learnable parameters (log scale) for different model \textbf{encoders}.  The dotted red line highlights the $80 \times 80$ pillars configuration used in our experiments reported in Table~\ref{tab:bp}.}
    \label{fig:flops} \vspace{-.5em}
\end{figure}

\begin{figure}[t]
    \centering
        \includegraphics[width=0.85\linewidth]{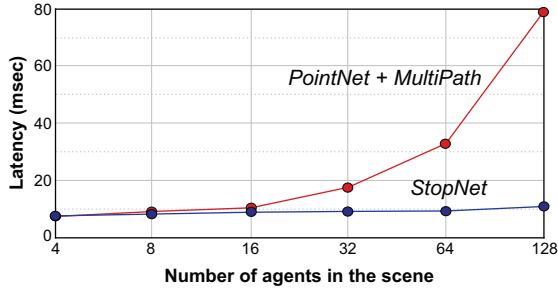}
    \caption{\ours scales well as the number of agents in the scene increase.  For agent-centric models, latency grows linearly with the number of agents.}
    \label{fig:latency} \vspace{-.5em}
\vspace{-10pt}
\end{figure}

\subsection{Metrics}
\myparagraph{Trajectory Metrics.}
We use two standard Euclidean-distance-based metrics~\cite{chang2019argoverse}: the minimum average displacement error, $\textnormal{minADE}_{k}$ = ${\underset{k}{\textnormal{min}}}\frac{1}{T}\sum_{t=1}^{T}||\mathbf{s}_t - \hat{\mathbf{s}}_t^k||_{2}$, and minimum final displacement error, $\textnormal{minFDE}_{k}$ = ${\underset{k}{\textnormal{min}}}||\mathbf{s}_T - \hat{\mathbf{s}}_T^k||_{2}$, where $\mathbf{s}$ denotes the ground-truth. We also report miss rate (MR), which measures the ratio of trajectories where none of the predictions are within $\{1, 2\}$ meters of the ground-truth according to FDE.

\myparagraph{Occupancy Metrics.}
Evaluation metrics for occupancy grids in the context of motion forecasting have not been well documented in the existing literature. An intuitive choice, however, is the mean cross entropy (CE) error between the predicted occupancy grids ${\mathcal{\hat{O}}}_{t}^{a}$ and the ground-truth $\mathcal{O}_{t}^{a}$ as $\frac{1}{WH}\sum_{x,y}\mathcal{H}(\mathcal{\hat{O}}^{a}_{t}, \mathcal{O}^{a}_{t})$. We also employ evaluation metrics commonly used for binary segmentation~\cite{chen2017deeplab}: We use a linearly-spaced set of thresholds in $[0, 1]$ to compute pairs of precision and recall values to estimate the area under the PR-curve as AUC. We also measure the probabilistic area of overlap as Soft Intersection-over-Union~\cite{mattyus2017deeproadmapper}:
\begin{equation}
    \small{\text{Soft-IoU} = \sum_{x,y}\mathcal{\hat{O}}^a_t\mathcal{O}^a_t /\left(\sum_{x,y}\mathcal{\hat{O}}^a_t + \sum_{x,y}\mathcal{O}^a_t - \sum_{x,y}\mathcal{\hat{O}}^a_t\mathcal{O}^a_t\right)}
\end{equation}

\subsection{Results}
\myparagraph{Trajectory Prediction.}
Table~\ref{tab:bp} compares our trajectory-only model $M_T$, and our co-trained model $M_{TO}$ with state-of-the-art trajectory prediction models, MultiPath~\cite{chai2019multipath}, TNT~\cite{zhao2020tnt}, and DESIRE~\cite{lee2017desire} on three different datasets.  To best evaluate the performance and latency characteristics of our encoder, we also compare our model with two agent-centric sparse encoders, namely VectorNet~\cite{gao2020vectornet}, and PointNet~\cite{qi2017pointnet} as used by CBP~\cite{tolstaya2021identifying}.  For an even comparison, we couple these agent-centric encoders with the same trajectory decoder~\cite{chai2019multipath} we have adapted in our architecture. Following existing work~\cite{gao2020vectornet, tolstaya2021identifying}, we compute per-agent embeddings of the world and concatenate it with per-agent state embeddings before feeding it to the trajectory decoder.

As Table~\ref{tab:bp} shows, our models match or exceed the performance of all the baselines, despite having a much smaller footprint. Note that the Argoverse dataset contains ground-truth future for a single agent, offering limited interactivity.  These results show the advantage of our sparse whole-scene encoder over existing raster and agent-centric methods.  Moreover, in all cases our co-trained model $M_{TO}$ achieves the best performance on all trajectory metrics.  This is likely due to the regularizing effect of unifying the two different output representation with a consistency loss.

{
\setlength{\tabcolsep}{4pt}
\renewcommand{\arraystretch}{1.3} 
\begin{table}[t]
    \caption{Trajectory prediction performance on different datasets. We report model performance on the validation set for Interaction and Argoverse datasets.}
    \vspace{-2em}
	\begin{center}
    	\resizebox{\linewidth}{!}{%
    	\begin{tabular}{@{}lccccc@{}} \toprule
    	    \textbf{Crowds dataset} & Sparse & Whole & minADE$_{6}\downarrow$ & minFDE$_{6}\downarrow$ & MR @ 1m, 2m \\\midrule
    	    MultiPath~\cite{chai2019multipath} &  & $\checkmark$ & 0.55 & 1.57 & 0.220, 0.385 \\
    	    VectorNet~\cite{gao2020vectornet} + MultiPath~\cite{chai2019multipath} & $\checkmark$ &  & 0.58 & 1.70 & 0.229, 0.399 \\
    	    PointNet~\cite{qi2017pointnet} + MultiPath~\cite{chai2019multipath} & $\checkmark$ &  & 0.53 & 1.60 & 0.235, 0.408\\
    	    $M_T$ (ours) & $\checkmark$ & $\checkmark$ & 0.51 & 1.54 & 0.223, 0.400 \\
    	    $M_{TO}$ (ours) & $\checkmark$ & $\checkmark$ & {\bf{0.51}} & {\bf{1.49}} & {\bf{0.215}}, {\bf{0.384}} \\\midrule\midrule
    	    \textbf{Interaction}~\cite{interactiondataset} & Sparse & Whole & minADE$_{6}\downarrow$ & minFDE$_{6}\downarrow$ & MR @ 1m, 2m $\downarrow$\\
    	    \midrule
    	    DESIRE~\cite{lee2017desire} &  & $\checkmark$ & 0.32 & 0.88 & - \\
    	    TNT~\cite{zhao2020tnt} & $\checkmark$ &  & 0.21 & 0.67 & - \\
    	    VectorNet~\cite{gao2020vectornet} + MultiPath~\cite{chai2019multipath} & $\checkmark$ &  & 0.30 & 0.99 & - \\
    	    $M_T$ (ours) & $\checkmark$ & $\checkmark$ & 0.21 & 0.60 & 0.150, 0.018 \\
    	    $M_{TO}$ (ours) & $\checkmark$ & $\checkmark$ & {\bf{0.20}} & {\bf{0.58}} & {\bf{0.136}}, {\bf{0.015}}\\\midrule\midrule
    	    \textbf{Argoverse}~\cite{chang2019argoverse} & Sparse & Whole & minADE$_{6}\downarrow$ & minFDE$_{6}\downarrow$ & MR @ 2m $\downarrow$ \\
    	    \midrule
    	    DESIRE~\cite{lee2017desire} &  & $\checkmark$ & 0.92 & 1.77 & 0.18\\
    	    VectorNet~\cite{gao2020vectornet} + MultiPath~\cite{chai2019multipath} & $\checkmark$ &  & {\bf0.80}  & 1.68 & {\bf 0.14} \\
    	    $M_T$ (ours)  & $\checkmark$ & $\checkmark$ & 0.87 & 1.68 & 0.19\\
    	    $M_{TO}$ (ours) & $\checkmark$ & $\checkmark$ & 0.83 & {\bf 1.54} & 0.19\\\bottomrule
        \end{tabular}
        }
     \end{center}
     \vspace{-2em}
    \label{tab:bp} 
\end{table}
}

\myparagraph{Scalability.}
Fig.~\ref{fig:flops} compares the number of flops and learnable parameters used by the \ours encoder vs. the whole-scene raster encoder from MultiPath and two agent-centric encoders.  Including the ResNet backbone, our nominal encoder with $80 \times 80$ pillars uses about $\sfrac{1}{10}$ the number of flops used by MultiPath.  Whole-scene approaches require a larger number of parameters as they need to have convolutional layers with a large receptive field.  However, our core encoder uses much fewer parameters.  Moreover, the compute required by our encoder is invariant to the number of agents---only a function of the pillar resolution.  Sparse encoders, on the other hand, require linearly more compute with growing number of agents.

Fig.~\ref{fig:latency} shows the latency of our model (encoder + decoder) as a function of the number of agents, compared with an agent-centric model. The variable latency of agent-centric models poses a problem for coordination of processes run by the AV. Note that raster representations also require rendering the model inputs, further increasing the effective latency.

\myparagraph{Occupancy Prediction.}
Table~\ref{tab:occ} shows occupancy prediction results on the Crowds dataset.  To evaluate our sparse input representation, we also train baseline models using BEV raster inputs.  Following existing work~\cite{bansal2018chauffeurnet, hong2019rules, chai2019multipath, phan2020covernet}, we render road structure, speed limits, traffic lights, and agent history at $400\times 400$ resolution and feed the stacked images to the model.  We also ablate the pillar resolution for our sparse encoder.  Results reflect the advantage of our sparse scene representation.  While $160$$\times$$160$ pillars work best, $80$$\times$$80$ pillars have comparable performance at lower complexity.

{
\setlength{\tabcolsep}{4pt}
\renewcommand{\arraystretch}{1.3} 
\begin{table}[t]
    \caption{Occupancy prediction results using raster / sparse inputs.
    }
    \vspace{-2em}
	\begin{center}
    	\resizebox{\linewidth}{!}{%
    	\begin{tabular}{@{}lcccccccccccc@{}} \toprule
    	    \multirow{4}{*}{Input} & \multirow{4}{*}{\# of pillars} & \multirow{4}{*}{CE $\downarrow$} & \multicolumn{4}{c}{Pedestrians} & \multicolumn{4}{c}{Vehicles} \\\cmidrule{4-11}
    	    & & & \multicolumn{2}{c}{AUC $\uparrow$} & \multicolumn{2}{c}{IoU $\uparrow$} & \multicolumn{2}{c}{AUC $\uparrow$} & \multicolumn{2}{c}{IoU $\uparrow$} \\\cmidrule{4-11}
    	    & & & 3s & 6s & 3s & 6s & 3s & 6s & 3s & 6s \\\midrule
    	    Raster & - & 19.2 & 0.48 & 0.24 & 0.21 & 0.13 & 0.84 & {\bf{0.73}} & 0.49 & 0.36\\ \midrule
    	    Sparse & 20$\times$20 & 18.6 & 0.48 & 0.24 & 0.22 & 0.13 & 0.83 & 0.71 & 0.50 & 0.35\\
    	    Sparse & 40$\times$40 & 17.6 & 0.54 & 0.26 & 0.25 & 0.14 & 0.86 & 0.72 & 0.50 & 0.36\\
    	    Sparse & 80$\times$80 & 17.2 & 0.56 & 0.27 & {\bf{0.27}} & {\bf{0.15}} & {\bf{0.87}} & {\bf{0.73}} & {\bf{0.53}} & {\bf{0.37}}\\
    	    Sparse & 160$\times$160 & {\bf{17.0}} & {\bf{0.59}} & {\bf{0.28}} & {\bf{0.27}} & {\bf{0.15}} & 0.86 & {\bf{0.73}} & {\bf{0.53}} & {\bf{0.37}}\\
    	    \bottomrule
        \end{tabular}}
     \end{center}
     \vspace{-2em}
    \label{tab:occ} 
\end{table}
}

\myparagraph{Occupancy Grids vs. Trajectories.}
Occupancy grid and trajectory representations have complementary advantages, which motivates \ours to support both output formats.  Trajectory models often output tens of potential trajectories per agent, which need to be taken into consideration as constraints in the planning algorithm.  The size of the trajectory outputs grows linearly with the number of agents in the scene, while the number of potential agent interactions grows quadratically.  This variability makes it challenging to complete planning for the AV under a fixed compute budget.  Occupancy grids require fixed compute to generate and consume regardless of the number agents in the scene.  They also capture the full extents of agent bodies, as opposed to just center locations, which simplifies calculating overlap probabilities.  On the other hand, trajectory sets can be represented as sparse sequences, which are more compact.  In scenes with few agents, processing few trajectories can be done faster than processing a dense probability map.

Fig.~\ref{fig:viz-examples} shows occupancy and trajectory predictions by our model on three sample urban driving scenes.  We observe that our occupancy representation is especially effective in situations where occupancy blobs can capture the collective behavior of groups, and eliminate the need for generating trajectory sets for individual agents.  The occupancy representation is particularly useful in busy urban scenes, where trajectory prediction models face challenges caused by noisy detection and poor tracking due to occlusions.

\begin{figure}[t]
    \centering
        \includegraphics[width=1.0\linewidth]{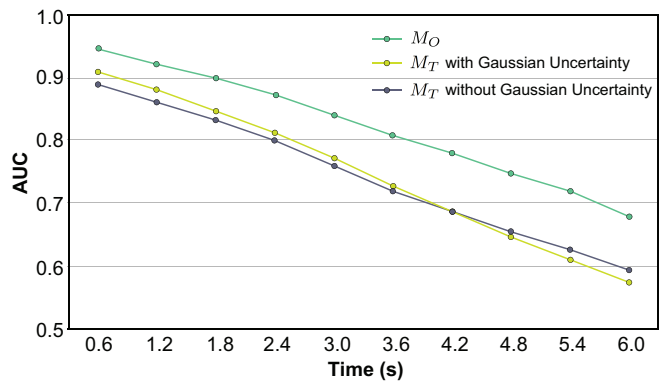}
    \caption{Comparison of our occupancy model $M_O$ with two versions of our trajectory model $M_T$ trained with and without Gaussian uncertainties on occupancy prediction for vehicles.  While $M_O$ predicts occupancy directly, the top six trajectory outputs from $M_T$ models have been converted (rendered) into an occupancy grid representation.  Results show that the rich non-parametric representation is more suitable for occupancy prediction.}
    \label{fig:occ-vs-traj} \vspace{-.5em}
\vspace{-10pt}
\end{figure}

Because of the different representations, it is difficult to directly compare the quality of trajectories with occupancy grids.  As a proxy, we convert predicted trajectories to occupancy by rendering agent boxes on locations predicted by the trajectory waypoints.  Since the model predicts multiple trajectories, we render each agent box with an intensity matching the associated probability for the corresponding trajectory.  Fig.~\ref{fig:occ-vs-traj} shows a comparison between our native occupancy model $M_O$ and occupancies converted from our trajectory model $M_T$.  We train two versions of $M_T$, once with and once without Gaussian uncertainties.  The two-dimensional variance of each Gaussian is factored in by first rendering the probability density function of the Gaussian and then convolving that with the rendered agent box.  As the plot shows, $M_T$ underperforms $M_O$ on this metric, which serves as validation for the utility of occupancy grids. Moreover, the plot shows that while including Gaussian uncertainties helps $M_T$ in the near future, it hurts performance over longer prediction horizons.  The position uncertainty of road agents is often more complex than a Gaussian mixture model, and is best represented with the rich non-parametric distributions supported by occupancy grids.

\section{Conclusions}
In this paper, we proposed \ours, a novel, efficient, and scalable motion forecasting method that accommodates sparse inputs in a whole-scene modeling framework, and co-trains trajectory and occupancy representations.  Our model has an almost fixed compute budget and latency, independent of the number of agents in the scene.  Likewise, our occupancy predictions can be consumed with fixed compute in a planning algorithm.  In addition to this higher efficiency and scalability, our experiments show that our model matches or outperforms performance of prior methods under standard trajectory and occupancy metrics. In future work, it would be interesting to extend the occupancy representation with per-pixel motion information, enabling the model to trace predicted occupancies back to the original agents.  Future research could explore applications of \ours to reasoning about occupancy of occluded objects---a challenging task for pure trajectory-based representations.

%\footnotesize{
%\myparagraph{Acknowledgments.}
%J. Kim is partially supported by the National Research Foundation of Korea grant (NRF-2021R1C1C1009608), Basic Science Research Program (NRF-2021R1A6A1A13044830), and ICT Creative Consilience program (IITP-2022-2022-0-01819).}

\bibliographystyle{IEEEtran}
\bibliography{bibliography}

\end{document}